\documentclass[10pt,twocolumn,letterpaper]{article}

\usepackage[pagenumbers]{wacv} 

\usepackage{tikz}
\usetikzlibrary{shapes.geometric, arrows.meta, positioning, calc}

\definecolor{wacvblue}{rgb}{0.21,0.49,0.74}
\usepackage[pagebackref,breaklinks,colorlinks,allcolors=wacvblue]{hyperref}

\def\wacvPaperID{1502} 
\def\confName{WACV}
\def\confYear{2027}

\title{Learning When to Listen: Gated Affect Fusion for Human Motion Prediction}

\author{Jingni Huang\\
University of Oxford\\
{\tt\small jingni.huang@kellogg.ox.ac.uk}\\
{\tt\small jingnih@gmail.com}
}

\begin{document}
\maketitle
\begin{abstract}
Human motion forecasting in unconstrained real-world videos remains challenging due to the ambiguity of future behaviors and the presence of noisy multimodal observations. While facial affect potentially provides complementary behavioral cues, its practical utility and mechanistic boundaries within motion forecasting frameworks remain poorly understood. In this work, we present a systematic study investigating the utility and temporal limitations of affect-conditioned forecasting in-the-wild. We establish a rigorous multimodal pipeline combining MediaPipe body pose trajectories with HSEmotion facial affect representations, and introduce the Gated Affect Transformer (GAT) to dynamically regulate cross-modal information flow. 

Through extensive multi-horizon evaluations under a strict subject-wise protocol, we demonstrate that naive early cross-modal concatenation consistently degrades forecasting accuracy relative to pose-only baselines. Conversely, our proposed gating mechanism stabilizes cross-modal integration by adaptively controlling the affective stream. Crucially, controlled counterfactual experiments using shuffled and randomized affect inputs reveal that the learned gate successfully suppresses unstructured cross-modal noise while remaining responsive to plausible affective signals. Furthermore, our empirical results indicate that facial affect features provide bounded, horizon-dependent predictive cues strictly within short-to-medium windows (e.g., 30 frames), whereas long-term trajectories remain predominantly governed by intrinsic kinematic continuity. Our findings provide empirical evidence that facial affect should be regarded as a complementary behavioral cue rather than a dominant driver of future motion, offering practical guidance for selective multimodal fusion in unconstrained human motion forecasting.
\end{abstract}
\section{Introduction}
\label{sec:introduction}

\begin{figure*}[t]
  \centering
  \includegraphics[width=0.98\textwidth]{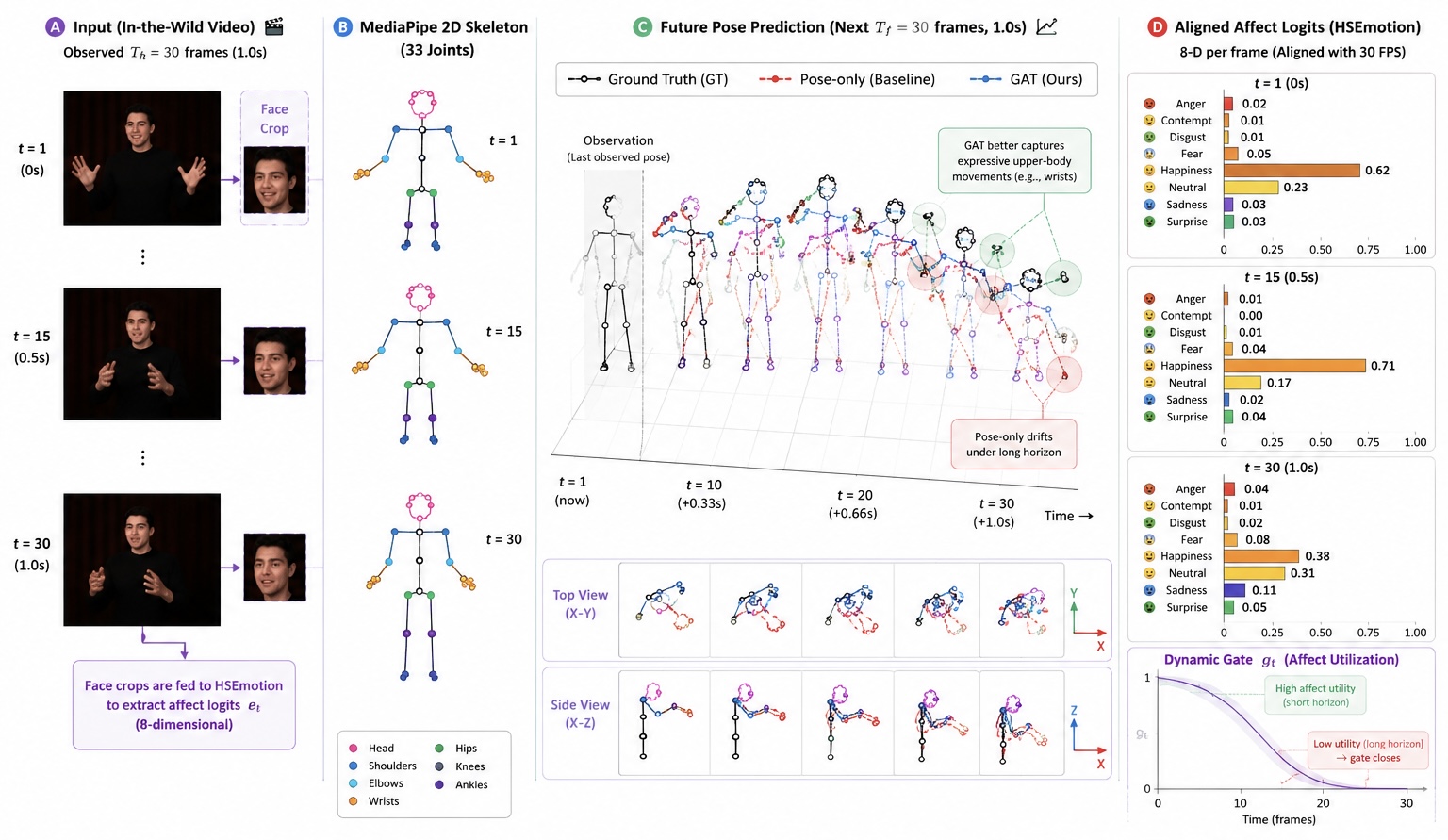}
  \caption{\textbf{Conceptual illustration of affect-conditioned human motion forecasting.} The figure illustrates the central hypothesis investigated in this work. Observed RGB frames are converted into synchronized body pose and facial affect representations, which are jointly used to predict future motion trajectories through an adaptive affect gating mechanism. This visualization is intended as a conceptual illustration rather than an experimental result. Quantitative and qualitative evaluations of this hypothesis are presented in Section~\ref{sec:experiments}}
  \label{fig:intro_concept}
\end{figure*}

Human motion forecasting is a fundamental problem in computer vision with applications in autonomous driving, human–robot interaction, and virtual environments. Existing forecasting methods predominantly rely on historical body kinematics to extrapolate future motion trajectories. However, human movement is also influenced by rich behavioral signals, including facial affect, which often reflects communicative intent and precedes expressive upper-body gestures. Whether such affective cues genuinely improve future motion prediction—and under what conditions they remain useful—remains largely unexplored.

Figure~\ref{fig:intro_concept} conceptually illustrates the central hypothesis investigated in this work. Rather than relying solely on observed body pose, we investigate whether synchronized facial affect provides complementary behavioral information for motion forecasting. However, integrating affect into a predictive pipeline is non-trivial. Facial expressions are inherently transient and noisy in unconstrained videos, and naive multimodal fusion can easily destabilize the learned kinematic representation, sometimes degrading performance even relative to pose-only baselines.

To address these limitations, we introduce the {Gated Affect Transformer (GAT)}, a novel multimodal framework designed to systematically study how continuous, frame-level facial affect representations can modulate future upper-body gestures through an adaptive gated fusion mechanism. Our approach leverages a dual-stream architecture: a kinematic stream utilizing normalized 2D coordinate sequences extracted via MediaPipe, and an affective stream capturing 8-dimensional emotion distribution vectors generated by HSEmotion. Instead of relying on rigid, hard-coded fusion boundaries, we design a learnable gating mechanism that continuously modulates the influence of the affective context based on its statistical reliability and predictive utility at any given timestamp. When facial expressions are ambiguous or corrupted, the gate automatically dampens the affect embeddings, safeguarding the model's structural tracking; conversely, during periods of highly expressive gesture-speech synchronization, the gate selectively amplifies the affective cues, adaptively balancing kinematic observations and facial affect cues during future motion prediction.

The principal contributions of this work are summarized as follows:
\begin{itemize}
    \item We present one of the first systematic studies of affect-conditioned human motion forecasting in unconstrained videos, characterizing when facial affect provides complementary predictive cues.
    \item We propose the Gated Affect Transformer (GAT), an architecture featuring an adaptive gating mechanism that dynamically balances kinematic history and multi-dimensional emotional signals to mitigate semantic trajectory drift.
    \item We conduct extensive quantitative and ablation studies to characterize when affect information is beneficial for motion forecasting. Our experiments show that facial affect serves as a complementary behavioral cue whose utility depends heavily on the forecasting horizon and the fusion strategy.
    \item We further provide qualitative visualizations illustrating representative prediction cases using real, shuffled, and random affect conditions, complementing the quantitative analyses.
\end{itemize}

\section{Related Work}
\label{sec:related_work}

\subsection{Human Motion Forecasting}
Generative human motion forecasting aims to predict future kinematic trajectories based on historical pose sequences. Early approaches primarily relied on recurrent neural networks (RNNs) and Long Short-Term Memory (LSTM) networks to model sequential progression~\cite{fragkiadaki2015recurrent,martinez2017human}. To better capture skeletal topology, Graph Convolutional Networks (GCNs) and trajectory dependency learning methods were subsequently introduced~\cite{yan2018spatial,mao2019learning}. Efficient MLP-based and diffusion-based motion models have also been explored for 3D human motion forecasting and generation~\cite{bouazizi2022motionmixer,tevet2023human}.

More recently, Transformer-based architectures have become dominant by leveraging self-attention to model long-range temporal dependencies~\cite{aksan2021spatio}.

However, these methods typically focus exclusively on kinematic continuity, ignoring non-verbal cues from other contextual modalities. As noted by Aksan et al., long-term motion prediction remains highly challenging because autoregressive prediction errors accumulate over time, leading to trajectory drift.

Existing multimodal forecasting methods generally assume that auxiliary modalities contribute consistently throughout the prediction horizon. However, relatively little attention has been paid to whether their utility varies over time or depends on forecasting horizon.

\subsection{Emotion-aware Human Behavior Modeling}
Recent facial expression recognition systems have achieved substantial progress on in-the-wild datasets such as AffectNet, enabling robust extraction of affective representations from unconstrained videos~\cite{mollahosseini2017affectnet}. Affective computing has long demonstrated that facial expressions, speech, and body language provide complementary behavioral cues for understanding human intent and social interaction~\cite{ekman1992argument,baltrusaitis2018openface}. These advances make it feasible to investigate whether facial affect can provide complementary behavioral information for downstream human motion forecasting.

In predictive frameworks, understanding facial emotional dynamics can yield critical contextual anchors regarding immediate emphasis or conversational transitions. However, in unconstrained videos, abstract facial expressions often exhibit high-frequency variations. Propagating raw affective tokens directly into temporal structural backbone layers can inject substantial cross-modal noise, necessitating a formal exploration of when and where emotion signals actually modulate geometric trajectories.

Our preliminary study~\cite{huang2026emotion} first suggested that facial affect may improve short-horizon human motion forecasting under controlled settings. In this work, we substantially extend that study through larger-scale in-the-wild experiments, systematic ablation analyses, and an investigation into the temporal utility and limitations of affect-conditioned forecasting.

\subsection{Multimodal Fusion Strategies}
Fusing heterogeneous modalities typically relies on early fusion via concatenation, late fusion via sequence pooling, or cross-attention alignments. While these strategies excel in stationary tasks, sequence-to-sequence forecasting presents unique temporal challenges. Existing multimodal fusion methods typically assume that all modalities contribute equally throughout the entire prediction horizon~\cite{baltrusaitis2019multimodal,tsai2019multimodal}. In contrast, we explicitly investigate whether affective cues should be dynamically regulated according to their temporal utility across varying future windows.

\section{Methodology}
\label{sec:methodology}

We study multimodal human motion forecasting from synchronized body pose and facial affect observations. Given an observed pose sequence $X_{1:T_h}=\{x_1,\dots,x_{T_h}\}$, where $x_t \in \mathbb{R}^{66}$ denotes the flattened 2D coordinates of 33 MediaPipe body landmarks, and an aligned affect sequence $A_{1:T_h}=\{a_1,\dots,a_{T_h}\}$, where $a_t \in \mathbb{R}^{8}$ denotes HSEmotion affect logits, the goal is to predict future pose coordinates $\hat{X}_{T_h+1:T_h+T_f}$ over a horizon $T_f$~\cite{lugaresi2019mediapipe}~\cite{hsemotion}. Unlike traditional motion forecasting methods that rely solely on kinematic continuity, our framework investigates whether facial affect provides complementary behavioral information for predicting future body motion. Rather than assuming that affect should always contribute equally, we explicitly model the dynamic importance of affect through a learnable gating mechanism.

\begin{figure}[t]
  \centering
  \resizebox{0.98\linewidth}{!}{
  \begin{tikzpicture}[
      node distance = 1.2cm and 0.4cm,
      box/.style = {rectangle, draw, fill=blue!5, minimum width=2.2cm, minimum height=0.6cm, align=center, rounded corners=2pt, font=\scriptsize},
      gate/.style = {circle, draw, fill=green!5, minimum size=0.6cm, font=\scriptsize},
      mult/.style = {circle, draw, fill=orange!10, minimum size=0.5cm, font=\scriptsize},
      add/.style = {circle, draw, fill=red!5, minimum size=0.5cm, font=\scriptsize},
      arrow/.style = {-{Stealth[scale=0.8]}, thick}
    ]

    \node (in_p) [align=center, font=\scriptsize] {Pose Input\\$X_t \in \mathbb{R}^{66}$};
    \node (in_a) [right=1.8cm of in_p, align=center, font=\scriptsize] {Affect Input\\$A_t \in \mathbb{R}^{8}$};

    \node (emb_p) [box, below=0.6cm of in_p] {Linear Projection\\$W_p X_t + b_p$};
    \node (emb_a) [box, below=0.6cm of in_a] {Linear Projection\\$W_a A_t + b_a$};

    \node (pt) [below=0.4cm of emb_p, font=\scriptsize] {$P_t \in \mathbb{R}^{128}$};
    \node (et) [below=0.4cm of emb_a, font=\scriptsize] {$E_t \in \mathbb{R}^{128}$};

    \node (gate_node) [box, fill=green!5, minimum width=1.6cm, below left=0.6cm and -0.4cm of et] {Gate $\sigma(W_g[P_t \parallel E_t]+b_g)$};
    
    \node (mult_p) [mult, below=1.8cm of pt] {$\times$};
    \node (mult_a) [mult, below=1.8cm of et] {$\times$};
    \node (node_1minus) [box, fill=orange!5, minimum width=1cm, left=0.3cm of mult_p] {$1 - g_t$};
    \node (add_fuse) [add, below left=0.8cm and 0.8cm of mult_a] {$+$};

    \node (trans) [box, fill=purple!5, minimum width=5cm, below=0.6cm of add_fuse] {2-Layer Transformer Encoder \\ (4 Attention Heads, $d=128$)};
    \node (head) [box, fill=gray!10, minimum width=4cm, below=0.5cm of trans] {Linear Prediction Head};
    \node (out) [below=0.5cm of head, align=center, font=\scriptsize\bfseries] {Predicted Future Poses\\$\hat{X}_{T_h+1:T_h+T_f}$};

    \draw [arrow] (in_p) -- (emb_p);
    \draw [arrow] (in_a) -- (emb_a);
    \draw [arrow] (emb_p) -- (pt);
    \draw [arrow] (emb_a) -- (et);
    
    \draw [arrow] (pt) |- ($(gate_node.west)-(0.1,0)$) |- (gate_node.west);
    \draw [arrow] (et) -- (gate_node.north);
    
    \draw [arrow] (gate_node.south) -- node[right, font=\tiny] {$g_t$} (mult_a);
    \draw [arrow] (gate_node.south) -| (node_1minus);
    \draw [arrow] (node_1minus) -- (mult_p);
    
    \draw [arrow] (pt) -- (mult_p);
    \draw [arrow] (et) -- (mult_a);
    
    \draw [arrow] (mult_p) |- (add_fuse);
    \draw [arrow] (mult_a) |- (add_fuse);
    
    \draw [arrow] (add_fuse) -- node[right, font=\tiny] {$H_t$} (trans);
    \draw [arrow] (trans) -- (head);
    \draw [arrow] (head) -- (out);

  \end{tikzpicture}
  }
  \caption{\textbf{System Architecture of Gated Affect Transformer (GAT).} Synchronized body poses ($X_t$) and facial affect features ($A_t$) are projected into a joint 128-dimensional latent space ($P_t, E_t$). A learnable scalar gate ($g_t$) dynamically interpolates between kinematics and affect representations, computing $H_t = (1-g_t)P_t + g_t E_t$ to adaptively regulate the affective stream before temporal modeling.}
  \label{fig:architecture}
\end{figure}
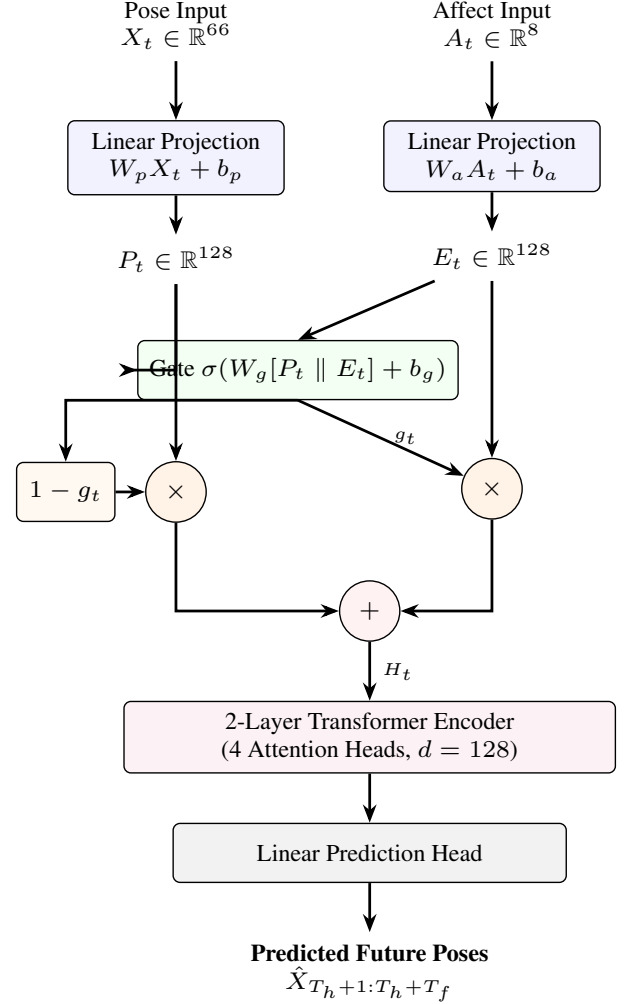

\subsection{Pose and Affect Embeddings}

We first project pose and affect features into a shared latent space:
\begin{equation}
P_t = W_p x_t + b_p,
\end{equation}
\begin{equation}
E_t = W_a a_t + b_a,
\end{equation}
where $P_t,E_t \in \mathbb{R}^{d}$ and $d=128$ in our experiments. The pose stream represents the dominant kinematic signal, while the affect stream provides a weak auxiliary behavioral cue.

\subsection{Gated Affect Fusion}

Naive early fusion can inject unstable cross-modal noise into the motion representation. To avoid forcing affective features into the prediction stream, we introduce a learnable scalar gate at each observed time step:
\begin{equation}
g_t = \sigma \left(W_g [P_t \parallel E_t] + b_g \right),
\end{equation}
where $g_t \in [0,1]$, $\sigma$ is the sigmoid function, and $[\cdot \parallel \cdot]$ denotes concatenation. The fused representation is then computed as:
\begin{equation}
H_t = (1-g_t)P_t + g_tE_t.
\end{equation}
This adaptive interpolation enables the model to regulate the contribution of affective information according to the observed multimodal context.

\subsection{Temporal Forecasting Backbone}

The fused sequence $H_{1:T_h}$ is processed by a two-layer Transformer encoder with four attention heads and hidden dimension 128~\cite{vaswani2017attention}. We use the final observed token representation to regress the future pose sequence:
\begin{equation}
\hat{X}_{T_h+1:T_h+T_f} = f_{\theta}(H_{1:T_h}),
\end{equation}
where $f_{\theta}$ denotes the Transformer encoder followed by a linear prediction head. We train the model using mean squared error over future normalized 2D joint coordinates. At evaluation time, we report MSE, MPJPE, ADE, and FDE over normalized coordinates\cite{ionescu2014human36m}.

\section{Experiments}
\label{sec:experiments}

\subsection{Dataset and Preprocessing}
We construct an in-the-wild multimodal motion forecasting dataset from public YouTube videos, including TED talks, vlogs, interviews, and public speaking scenarios. For each subject, we extract three clips of approximately 15–25 seconds. The dataset contains naturally occurring conversational behaviors rather than scripted actions.

Body pose is extracted using MediaPipe Pose with 33 landmarks, retaining only normalized 2D coordinates $(x,y)$, resulting in a 66-dimensional pose vector per frame. Facial affect representations are extracted using the pre-trained HSEmotion model trained on AffectNet~\cite{mollahosseini2017affectnet}. Pose and affect features are temporally aligned at the frame level at 30 frames per second (fps).

Unlike laboratory motion capture datasets, our videos exhibit substantial appearance diversity, including viewpoint changes, partial occlusions, and spontaneous conversational behaviors, making the benchmark considerably more challenging.

\subsection{Evaluation Protocol}
We use a strict subject-wise split to avoid identity leakage, ensuring that individuals in the test set never appear during training. After preprocessing and alignment, the final usable index contains 243 clips, with 169 allocated for training, 39 for validation, and 35 for testing. Models observe a history sequence of $T_h = 30$ frames (1 second) and predict future horizons of $T_f \in \{15, 30, 60, 90\}$ frames, corresponding to 0.5s, 1.0s, 2.0s, and 3.0s future windows.

\subsection{Implementation Details}
All models are implemented in PyTorch and trained with the Adam optimizer using a learning rate of $10^{-3}$~\cite{kingma2015adam}. We use a batch size of 64 and train each model for 10 epochs. For each prediction horizon, the checkpoint with the lowest validation MPJPE is selected for final test evaluation. The Transformer backbone uses two encoder layers, four attention heads, a feed-forward dimension of 256, and a latent dimension of 128. All reported gated models are trained with a fixed random seed of 42 for reproducibility.

\subsection{Baselines}
To benchmark our proposed Gated Affect Transformer (GAT), we compare against the following baseline configurations:
\begin{itemize}
    \item \textbf{Pose-only Transformer:} A single-stream temporal backbone that models kinematics exclusively from the 66-dimensional landmarks, ignoring affective cues.
    \item \textbf{Concat Transformer:} An early-fusion baseline where the 66-dimensional pose sequence and the aligned affect stream, temporally averaged into an 8-dimensional context vector, are concatenated before entering the Transformer backbone. Unlike the proposed Gated Affect Transformer, this baseline discards temporal affect dynamics by averaging affect features over the observation window.
\end{itemize}

\subsection{Metrics}
We evaluate forecasting accuracy using four standard geometric metrics computed over the normalized 2D joint coordinates: Mean Squared Error (\textbf{MSE}), Mean Per-Joint Position Error (\textbf{MPJPE}), Average Displacement Error (\textbf{ADE}), and Final Displacement Error (\textbf{FDE}). All displacement metrics are reported as normalized distance units per joint.

\subsection{Main Results}
Table~\ref{tab:main_results} compares the performance of the pose-only prediction, naive affect concatenation, and our proposed gated affect fusion across multiple prediction horizons.

The results show that affective information is not universally beneficial for geometric accuracy. Naive concatenation (Concat Transformer) exhibits substantial degradation in performance at short-to-medium horizons (15f and 30f) compared to the pose-only baseline. In contrast, our Gated Affect Transformer preserves short-to-medium horizon accuracy and achieves slight improvements at 30f, demonstrating that the gating mechanism provides more stable cross-modal integration during early prediction windows. However, at longer horizons (60f and 90f), the pose-only Transformer remains the strongest model, indicating that long-term trajectory generation is primarily governed by kinematic continuity.

\begin{table}[t]
\centering
\caption{Main forecasting results across prediction horizons. We report normalized MPJPE and FDE; lower is better. Bold values indicate the top performance within each horizon.}
\label{tab:main_results}
\small
\begin{tabular}{lcccc}
\hline
\textbf{Method} & \textbf{15f} & \textbf{30f} & \textbf{60f} & \textbf{90f} \\
\hline
\multicolumn{5}{c}{\textit{MPJPE}} \\
Pose-only Transformer & 0.0619 & 0.0730 & \textbf{0.0808} & \textbf{0.0873} \\
Concat Transformer & 0.0711 & 0.0910 & 0.0873 & 0.0929 \\
Gated Affect Transformer & \textbf{0.0612} & \textbf{0.0725} & 0.0878 & 0.0957 \\
\hline
\multicolumn{5}{c}{\textit{FDE}} \\
Pose-only Transformer & 0.0704 & 0.0862 & \textbf{0.0959} & \textbf{0.1035} \\
Concat Transformer & 0.0787 & 0.1026 & 0.1027 & 0.1086 \\
Gated Affect Transformer & \textbf{0.0703} & \textbf{0.0840} & 0.0977 & 0.1110 \\
\hline
\end{tabular}
\end{table}

\subsection{Affect Corruption Study}
To examine whether the gating mechanism responds to structured affective signals rather than merely digesting additional input dimensions, we evaluate the learned gated architecture under two corrupted control streams:
\begin{itemize}
    \item \textbf{Shuffled Affect:} Replaces the aligned affect sequence with an affect sequence sampled from a different video clip within the same data split.
    \item \textbf{Random Affect:} Replaces the affect stream with deterministic Gaussian noise ($\mu=0, \sigma=1$).
\end{itemize}

As shown in Table~\ref{tab:affect_corruption}, the model utilizing real aligned affect shows its targeted advantage at the 30f horizon. Crucially, the empirical analysis of the gating values reveals that when random noise is injected, the learned gate mean drops sharply from 11--13\% to 3--6\%. This indicates that the learned gate assigns a conservative but non-zero weight to affective information while suppressing unreliable affect signals.

\begin{table}[t]
\centering
\caption{Affect corruption study using the gated architecture. Gating behavior demonstrates active suppression of random cross-modal noise.}
\label{tab:affect_corruption}
\small
\begin{tabular}{lcccc}
\hline
\textbf{Condition} & \textbf{15f} & \textbf{30f} & \textbf{60f} & \textbf{90f} \\
\hline
\multicolumn{5}{c}{\textit{MPJPE}} \\
Real Affect & 0.0612 & \textbf{0.0725} & 0.0878 & 0.0957 \\
Shuffled Affect & \textbf{0.0592} & 0.0789 & \textbf{0.0873} & \textbf{0.0952} \\
Random Affect & 0.0598 & 0.0786 & 0.0902 & 0.0960 \\
\hline
\multicolumn{5}{c}{\textit{Gate Mean}} \\
Real Affect & 0.1199 & 0.1181 & 0.1369 & 0.1347 \\
Shuffled Affect & 0.1125 & 0.1118 & 0.1329 & 0.1345 \\
Random Affect & 0.0602 & 0.0342 & 0.0501 & 0.0490 \\
\hline
\end{tabular}
\end{table}

\subsection{Qualitative Case Study}
We visualize a representative 30-frame test case by comparing predicted joint trajectories under real, shuffled, and random affect inputs. This case study is intended as a qualitative diagnostic rather than standalone evidence of superior accuracy. 

\begin{figure}[t]
  \centering
  \includegraphics[width=\linewidth]{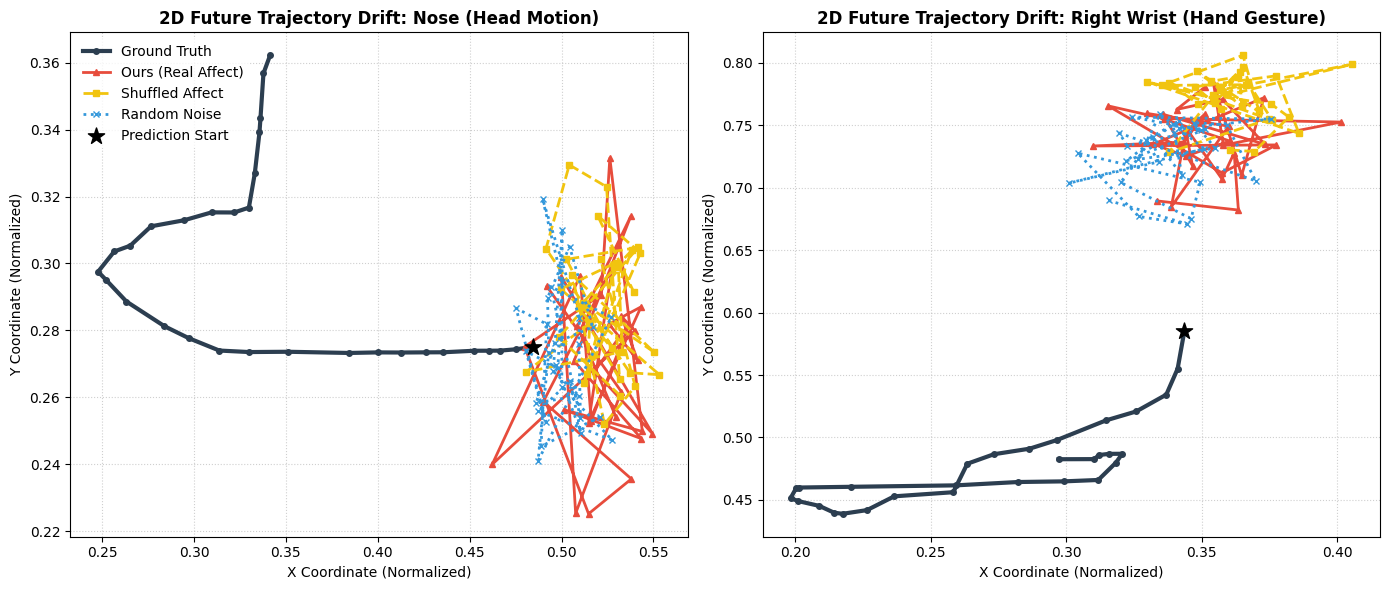}
  \caption{\textbf{Qualitative Trajectory Analysis (Experiment 7).} Comparison of predicted 30-frame future joint trajectories under real, shuffled, and random Gaussian affect conditions. Different affect conditions lead to observable trajectory variations, particularly for high-variance upper-limb joints, illustrating that affective inputs influence the predicted motion distribution. These qualitative differences are consistent with the quantitative findings that affect acts as a bounded behavioral modifier rather than a dominant predictor of future motion.}
  \label{fig:trajectory_case}
\end{figure}
Overall, the quantitative and qualitative analyses consistently indicate that affective information provides useful but limited complementary cues, primarily benefiting short-term motion forecasting rather than long-term trajectory generation.
  

Although this single example does not constitute quantitative evidence, the observed trajectory variations are consistent with the quantitative findings reported in Table II, suggesting that affective information influences short-term motion prediction in a bounded manner. However, the case-specific MPJPE and FDE vary across conditions, reinforcing our quantitative finding that affect provides bounded and horizon-dependent cues rather than universally improving geometric accuracy.

\section{Discussion and Limitations}
\label{sec:discussion}
The experimental results reveal an interesting observation:
affective information improves motion prediction only within a limited temporal horizon.
\subsection{The Horizon-Dependent Impact of Affect}
Our empirical findings indicate that facial affect expressions provide bounded, horizon-dependent cues rather than global geometric regularizations. The affective signal delivers its clearest advantage specifically at the medium-short horizon ($30$f / 1.0s). This observation is consistent with previous motion forecasting studies showing that prediction uncertainty increases rapidly as temporal horizons extend, making long-range motion primarily governed by kinematic continuity rather than short-lived contextual cues~\cite{martinez2017human,mao2019learning,aksan2021spatio}.

This localized benefit aligns with human behavioral dynamics: facial expressions typically manifest as transient, short-lived emotional bursts (e.g., micro-expressions or rapid conversational shifts) that modulate immediate upper-body kinematics. Over a 1-second window, these emotional cues provide predictive context regarding conversational intent or immediate gesture emphasis. 

\subsection{Kinematic Dominance in Long-Term Forecasting}
At extended forecasting horizons ($60$f and $90$f), the pose-only baseline consistently outperforms all affect-infused variants. This performance inversion suggests that long-term motion prediction is heavily dominated by intrinsic kinematic continuity and physical constraints (e.g., momentum, skeletal joint limits, and gravity)~\cite{martinez2017human,mao2019learning,aksan2021spatio}. 

As the temporal horizon expands, the predictive power of a transient facial expression decays exponentially, and forcing the network to attend to the affect stream introduces cross-modal noise that disrupts long-term trajectory generation. Our gating mechanism reflects this physical reality by maintaining a conservative allocation, yet the residual signaling still degrades long-range spatial geometry compared to an isolated kinematic backbone.

\subsection{Limitations}
While this work offers insights into multimodal behavioral fusion, several key limitations remain:
\begin{itemize}
    \item \textbf{2D Spatial Constraints:} Our framework intentionally restricts its experiments to normalized 2D coordinates. We adopt this boundary because the pseudo-depth ($Z$-coordinate) estimated by monocular MediaPipe is considerably noisier than the image-plane coordinates in unconstrained, in-the-wild YouTube videos, which would introduce uncontrolled tracking artifact errors into the forecasting baseline.

    \item \textbf{In-the-Wild Dataset Noise:} The YouTube dataset captures complex, unconstrained scenarios (TED talks, interviews, and vlogs). While providing high ecological validity, it features frequent camera cuts, partial body occlusions, and diverse framing boundaries that introduce structural noise into the sequence tracking.
    \item \textbf{Upstream Extractor Dependency:} The affect representations are entirely dependent on the pre-trained HSEmotion backend. Any systematic classification errors or domain shifts in the facial emotion logit extractor propagate directly into our gating module.
    \item \textbf{Qualitative Nature of Interventions:} Our counterfactual analysis remains an isolated qualitative case study. While it highlights model sensitivity to changed affect tokens, a comprehensive causal verification would require a large-scale, automated counterfactual generation pipeline.
\end{itemize}

\section{Conclusion}
\label{sec:conclusion}

In this work, we presented the Gated Affect Transformer (GAT) to systematically analyze the impact of facial affect on multimodal human motion forecasting. Rather than relying on naive concatenation, our architecture introduces a learnable scalar gate to dynamically weigh kinematic features against aligned emotional logits. 

Our empirical evaluations reveal that facial affect provides bounded, horizon-dependent predictive cues. While gated affect fusion yields measurable improvements at medium-short horizons (30f / 1.0s), affective signals become less informative at longer horizons (60f and 90f), where motion is dominated by intrinsic kinematic continuity. Corruption studies further verify that the learned gate suppresses random cross-modal noise while remaining responsive to structured affective signals.
Future work will build upon our previous short-horizon affect-conditioned motion forecasting framework~\cite{huang2026emotion} by investigating richer affect representations and longer temporal reasoning. We also plan to incorporate additional behavioral modalities, including speech and textual context, extend the framework to robust 3D human motion forecasting, and develop large-scale automated counterfactual evaluation pipelines. More broadly, we will investigate whether affect-conditioned motion forecasting can be extended toward human-centric predictive world models, where physical dynamics, facial affect, speech, and contextual signals are jointly modeled for anticipating future human behavior, following recent advances in latent predictive representation learning and world modeling~\cite{lecun2022path,assran2023self}.

We believe that selectively integrating behavioral signals with physical dynamics provides a promising direction toward behavior-aware human motion forecasting and, more broadly, future human-centric predictive world models.
{
    \small
    \bibliographystyle{ieeenat_fullname}
    \bibliography{main}
}

\end{document}